\documentclass[letterpaper, 10 pt, journal, twoside]{IEEEtran}
\usepackage{amsmath,amsfonts}
\usepackage{algorithmic}
\usepackage{algorithm}
\usepackage{array}
\usepackage[caption=false,font=normalsize,labelfont=sf,textfont=sf]{subfig}
\usepackage{textcomp}
\usepackage{stfloats}
\usepackage{url}
\usepackage{verbatim}
\usepackage{graphicx}
\usepackage{cite}
\usepackage[capitalize]{cleveref}
\usepackage{multicol}
\usepackage{multirow}
\usepackage{threeparttable}
\usepackage{booktabs}
\usepackage{verbatim}
\usepackage[export]{adjustbox}
\usepackage{makecell}

\begin{document}

\makeatletter
\def\ps@IEEEtitlepagestyle{%
  \def\@oddfoot{\mycopyrightnotice}%
  \def\@oddhead{\hbox{}\@IEEEheaderstyle\leftmark\hfil\thepage}\relax
  \def\@evenhead{\@IEEEheaderstyle\thepage\hfil\leftmark\hbox{}}\relax
  \def\@evenfoot{}%
}
\def\mycopyrightnotice{%
  \begin{minipage}{\textwidth}
  \centering \scriptsize
  Copyright 2377-3766 ~\copyright~2023 IEEE. Personal use of this material is permitted. Permission from IEEE must be obtained for all other uses, in any current or future media, including\\reprinting/republishing this material for advertising or promotional purposes, creating new collective works, for resale or redistribution to servers or lists, or reuse of any copyrighted component of this work in other works by sending a request to pubs-permissions@ieee.org.
  \end{minipage}
}
\makeatother

\title{
MS-Net: A Multi-Path Sparse Model for Motion Prediction in Multi-Scenes
}

\author{Xiaqiang Tang$^{* 1}$, Weigao Sun$^{* 2}$, Siyuan Hu$^{1}$, Yiyang Sun$^{1}$ and Yafeng Guo$^{\dagger 1}$
\thanks{ACCEPTED BY IEEE ROBOTICS AND AUTOMATION LETTERS, VOL. 9, NO. 1, JANUARY 2024}
\thanks{This research was supported by the National Natural Science Foundation of China under Grants 62373284 and 61973239.}
\thanks{$*$ Indicates equal contribution.}
\thanks{$\dagger$ Indicates the corresponding author.}
\thanks{$^1$Xiaqiang Tang, Siyuan Hu, Yiyang Sun, and Yafeng Guo are with the Department of Control Science and Engineering, Tongji University, Shanghai, China
{\tt\footnotesize (e-mail: xiaqiangtang@tongji.edu.cn;2130734@tongji.edu.cn;
2111123@tongji.edu.cn;yfguo@tongji.edu.cn)}}
\thanks{$^2$Weigao Sun is with the Shanghai Artificial Intelligence Laboratory, Shanghai, China 
{\tt\footnotesize (e-mail: sunweigao@outlook.com)}}
\thanks{Digital Object Identifier (DOI): 10.1109/LRA.2023.3338414}
}

\maketitle

\begin{abstract}

   The multi-modality and stochastic characteristics of human behavior make motion prediction a highly challenging task, which is critical for autonomous driving. While deep learning approaches have demonstrated their great potential in this area, it still remains unsolved to establish a connection between multiple driving scenes (e.g., merging, roundabout, intersection) and the design of deep learning models. Current learning-based methods typically use one unified model to predict trajectories in different scenarios, which may result in sub-optimal results for one individual scene. 
    To address this issue, we propose \textbf{M}ulti-\textbf{S}cenes Network (aka. MS-Net), which is a multi-path sparse model trained by an evolutionary process. MS-Net selectively activates a subset of its parameters during the inference stage to produce prediction results for each scene. In the training stage, the motion prediction task under differentiated scenes is abstracted as a multi-task learning problem, an evolutionary algorithm is designed to encourage the network search of the optimal parameters for each scene while sharing common knowledge between different scenes. Our experiment results show that with substantially reduced parameters, MS-Net outperforms existing state-of-the-art methods on well-established pedestrian motion prediction datasets, e.g., ETH and UCY, and ranks the 2nd place on the INTERACTION challenge.
\end{abstract}


\begin{figure}[ht!]
    \centering
   \includegraphics[width=0.9\linewidth]{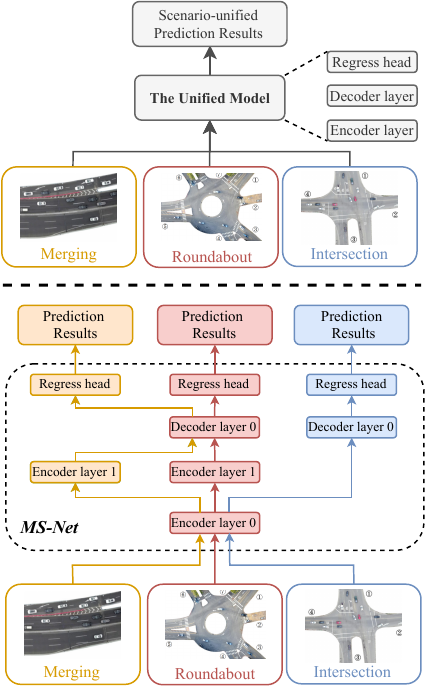}
   \caption{\textbf{A comparison between the current unified motion prediction model (top) and our MS-Net (down).} Note that the model structure in the figure is illustrative and doesn't reflect our actual experimental setup.}
   \label{fig_logic}
\end{figure}

\section{Introduction}
\subsection{Overview}

    \IEEEPARstart{F}{or} autonomous driving, precise prediction of future states for various road users, such as vehicles, cyclists, and pedestrians, is essential to ensure safe, comfortable, and human-like driving behavior. This paper centers on the development of a multi-path sparse model tailored to address diverse traffic scenarios within the motion forecasting framework. Our objective is to introduce a new methodology that improves motion prediction performance by leveraging shared information across different scenes, retaining distinctive scenario-specific knowledge, and optimizing the parameter efficiency of the model based on scenario complexity.

    Existing learning-based methods for behavior prediction, such as \cite{1,2,3,4,6,7}, primarily rely on a unified model for predicting the trajectories from various scenes. Such methods lack the step of abstracting different scenarios in the modeling process and suffer from model collapse problems \cite{7,mmtrans,multipath}. In contrast, rule-based multi-scenario learning has been widely adopted in the industrial field, such as the trajectory prediction module on Apollo \cite{baiduapollo}, which uses a hybrid structure that switches between simple models (e.g., Kalman filter \cite{filter}) and complex learning-based models (e.g., VectorNet \cite{6}) for highly interactive traffic scenario. While these methods incorporate scene extraction in the modeling of complex scenarios, the problem remains that common knowledge among different scenarios cannot be shared, resulting in a significant waste of computing resources and time during model training. Hence, a pertinent question emerges: \textit{Is it possible to design a method that dynamically adapts the network architecture according to scenario complexity while concurrently facilitating knowledge sharing across diverse scenarios?}

    In order to overcome the limitations associated with current behavior prediction methods, it is worthwhile to explore alternative approaches that have demonstrated success in other domains. In the fields of natural language processing as well as computer vision, sparse models such as mixture-of-experts (MoE) models \cite{nlpMoe} have been demonstrated to outperform unified models. This has prompted us to investigate whether a sparse model can be utilized to leverage different experts or network paths for motion prediction in diverse traffic scenarios.

    However, it is difficult to establish a connection between the topology of the traffic scene and the network design for deep learning, so unlike designing different experts for different tasks, we propose to decompose different traffic scenes into groups and employ a self-evolutionary approach to learn a sparse model that activates a subset of the parameters in the network for trajectory prediction in different scenarios. We name this framework \textbf{\textit{MS-Net}} (\textbf{\textit{M}}ulti-\textbf{\textit{S}}cenes \textbf{\textit{Net}}work). As depicted in \cref{fig_logic} (Top): existing methods rely on a unified model to predict trajectories for various scenarios; (Bottom): MS-Net employs a multi-path sparse model that only activates a small subset of parameters in the model for each traffic scenario (e.g., merging, roundabout, intersection) in the inference stage.
    
\subsection{Related Work}

\begin{figure*}[ht!]
     \centering
     \includegraphics[width=\textwidth]{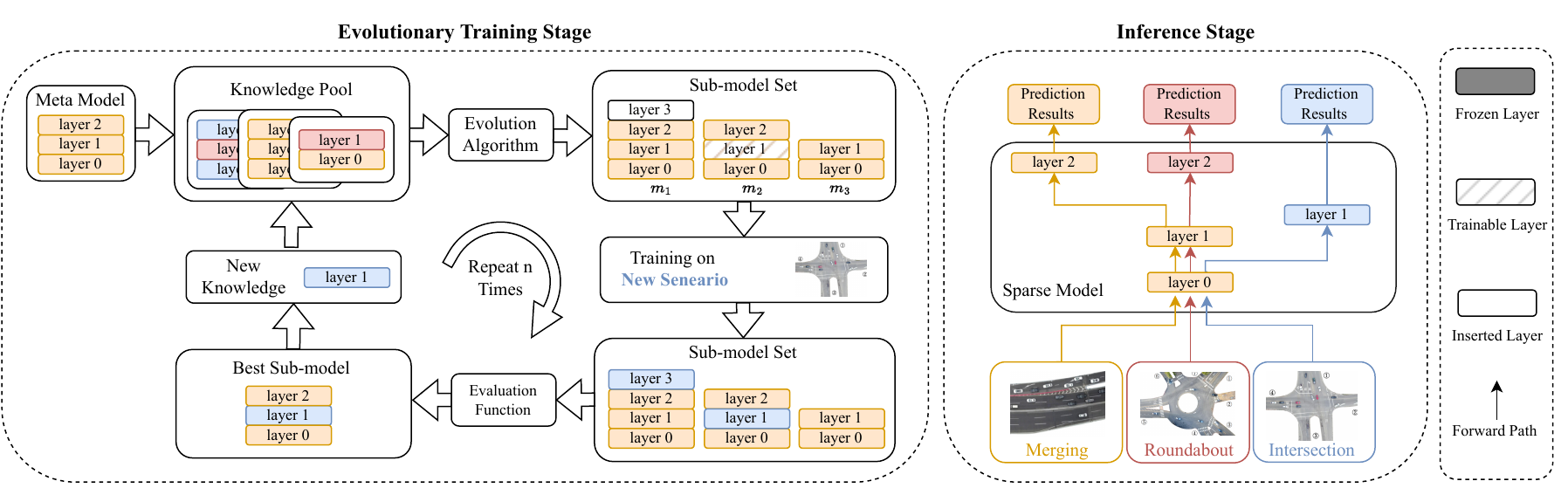}
     \caption{\textbf{The overall training and inference processes of MS-Net.} We choose a meta-model to initialize the Knowledge Pool. For new scenarios, a template model is selected from the pool, and sub-models are generated using an evolutionary algorithm. These sub-models are trained for scenario-specific knowledge. An evaluation function balances accuracy with parameter count. Since most parameters are inherited, we only need to add additional parameters (i.e., new knowledge) to the Knowledge Pool. In the inference stage, we form a sparse model from the Knowledge Pool, activating only a small part of all parameters for each scenario to achieve a scenario-distinct motion prediction model.}
     \label{method}
\end{figure*}

Currently, the learning-based trajectory prediction methods are mainly based on the unified model which applies one model for all scenarios. LaneGCN\cite{2} combines convolutional and graph neural networks to extract the interactive semantics and fuse information among road topology, traffic rules, and agents. Multipath++\cite{3} uses the output of anchor embeddings from each feature as the input and two heads respectively to predict the probability of intentions and tracks. The mask strategies to filter features in different scenarios are used in Scene Transformer\cite{4}. Future tracks of the influencers are added to the features M2I\cite{5}. VectorNet\cite{6} proposes a multi-level graph and the vector format to unify the coding of road structures and agents, thus solving the information fusion problem between context and agents.

However, there are several common issues with the unified model method. Using the same model structure in different scenarios is inefficient and may result in sub-optimal performance. \cite{sanchez2022scenario} illustrates how a unified model can yield erroneous conclusions in certain prediction scenes, \cite{huang2022tip} demonstrates that existing prediction networks fail to distinguish between scenarios, resulting in sub-optimal performance in subdivided prediction scenes, it incorporates downstream task losses into the prediction model. Additionally, model collapse is challenging due to insufficient scenario extraction.
\cite{multipath,7} deal with the model-collapse problem by transforming intention recognition into a classified task based on anchor trajectories or intention points. \cite{Kim2022DiverseMT, Park2020DiverseAA} apply auxiliary loss (e.g. symmetric cross-entropy and Lane loss respectively) to encourage the model to make diversified predictions that cover all plausible modes in the future trajectory. 
such methods\cite{multipath,7} need to transform intention recognition into classified task supervision or use additional methods\cite{mmtrans,autobot} to ensure the diversity of the prediction result.

Inspired by previous works our method goes one step further by decoupling prediction into different scenes and employing a multi-path sparse model. We argue that a scenario-distinct model can generate superior results compared to the scenario-unified model.

To design an efficient network structure, methods like reinforcement learning control network depth or layer width \cite{efficientnas}. UberNet \cite{9} introduced hard-parameter sharing for varied vision tasks. Soft-parameter sharing emerged in deep MTL through cross-stitch networks \cite{10}. Sluice networks \cite{11} expanded on this by selectively sharing skip connections, and MTAN \cite{12} employed a shared task-specific backbone. PAP-Net \cite{13} utilized recursive extraction for task-specific patterns, while \cite{15} updated shared network weights based on task-specific gradients. Recent studies have also explored the use of evolutionary methods to search for better Transformer architectures, such as performing an evolutionary search for SubTransformers with weight sharing in \cite{Hardware_Aware_Transformers}.

As the complexity of traffic scenarios varies, the difficulty of motion prediction also differs. To avoid manually designing each path of the model, inspired by recent research\cite{evolved_transformer, Hardware_Aware_Transformers}, we propose using the evolutionary method to control the knowledge sharing between paths and the growth of paths' depth for each prediction scenario.

\subsection{Contribution}

Specifically, targeting motion forecasting networks, we design four evolutionary methods to optimize the model for the current scenario. Firstly, the Model Evolution mechanism compresses or expands deep neural network models by randomly removing or adding layers during the generation of sub-models. Secondly, the Knowledge Transfer mechanism involves the sub-model creating copies of selected layers from the parent model, which can be fine-tuned during training, while the remaining portion of the parent model layer is shared in a frozen state to prevent catastrophic forgetting\cite{2017overcomingforgetting,hadsell2020embracing,hayes2020remind}. This mechanism leverages knowledge acquired from other scenes and optimizes it for its own scene. Thirdly, the Scoring Function selects the optimal sub-model with the lowest number of additional parameters and the highest accuracy from candidate sub-models. Lastly, Hyperparameter Tuning is achieved using the random walk algorithm to tune the hyperparameters of sub-models, which further optimizes the learning procedure for the sub-model.

The proposed MS-Net not only improves forecasting accuracy by reducing uncertainty and diversity across different modalities within a single scene, but it also achieves this with a significantly reduced effective parameter count. Moreover, MS-Net exhibits the capability to adaptively evolve its self-expression abilities, dynamically adjusting the number of network layers based on the scene complexity. We evaluate our method on multiple multi-scene datasets, e.g., ETH\cite{21}, UCY\cite{22} and INTERACTION datasets\cite{INTERACTIONDATASET}, results show that our method effectively boosts the performance of existing state-of-the-art approaches and ranks the 2nd place on the INTERACTION Challenge.

Our primary contributions can be summarized as follows:

\begin{itemize}
\item To the extent of our current knowledge, we present a pioneering approach by introducing a multi-path sparse model to address the task of motion prediction in autonomous driving, thereby framing motion prediction as a multi-task learning challenge.

\item Given the wide spectrum of diverse scenarios encountered in motion prediction, we propose to harness the power of evolutionary learning techniques to complete the training of a multi-path sparse model \textbf{\textit{MS-Net}}. This method aims at not only augmenting the overall predictive performance on multi-scenes but also enhancing parameter efficiency.

\item Our MS-Net demonstrates its performance as well as computation efficiency across a spectrum of real-world datasets encompassing seven distinct scenarios. The inherent generalizability of MS-Net suggests that our methodology holds promise for potential integration into a multitude of existing motion prediction models.

\end{itemize}

\section{Method}
In this section, we formally introduce MS-Net, whose pipeline is depicted in \cref{method}.
MS-Net initializes with a single meta-model and employs an evolutionary algorithm to search for optimal sub-models to form a sparse model specific to the traffic scenario at hand. This process involves three stages: 1) inheriting knowledge from the meta-model; 2) evolution of the sub-model; 3) fine-tuning the hyperparameters to achieve the best performance. The pseudo-code for training MS-Net is detailed in \cref{alg3}. Our evolutionary algorithm has the potential to be applied to a wide range of existing trajectory prediction networks, therefore it can capture recent advancements in motion forecasting.

\subsection{Meta Model}  

    Our algorithm starts with a meta-model (i.e., parent model), we choose popular open-source unified motion forecasting networks as the meta-model to provide a solid foundation for our algorithm. These meta-models are subsequently fine-tuned using evolutionary techniques to achieve optimal performance in a specific transportation scenario. Specifically, we have selected the original Transformer Network (TF)\cite{predictionTransfomer}, Agent-Aware Transformer\cite{AgentFormer}, and AutoBot\cite{autobot} as the meta-model. These meta-models were chosen based on their demonstrated effectiveness in prior research.
        
 \begin{figure*}[ht!]
 \centering
     \includegraphics[width=0.9\textwidth]{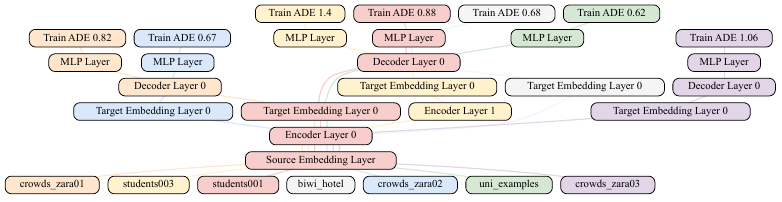}
     \caption{\textbf{The MS-Net structure diagram obtained by training on ETH/UCY.} We use seven training sets from ETH/UCY as the seven separate scenarios, and take the model from \cite{predictionTransfomer} as the meta-model. The modules with the same color in the figure indicate the network layers obtained from the same scenario, and it can be seen that the layers such as the encoder, and embedding layers are commonly reused by each task. In more complex scenarios, such as the student003 dataset, the network adaptively adds "Encoder Layer 1" for better handling such scenarios.}
     \label{fig3}
\end{figure*}

 \subsection{Knowledge Pool}  
 
    The Knowledge Pool serves as a repository for shared knowledge and is initialized with a single meta-model from an existing motion forecasting network. During the evolution process, a model is chosen from the Knowledge Pool as a template (i.e. parent model), and a set of modifications are applied to generate a group of sub-models (i.e. child model). After training these on a scene, the best-scoring sub-model's knowledge is added to the Knowledge Pool.

    The model selection follows the strategy outlined in \cref{parent-select}, whereby a higher-scoring model that has generated fewer sub-models is preferred. If the higher-scoring models are unable to generate improvements, the algorithm automatically transitions toward a more exploratory behavior.

\begin{algorithm}[t]
\caption{MS-Net Training Algorithm}
\label{alg3}
\begin{algorithmic}[1]
\STATE \textbf{Input:} The Knowledge Pool set $\mathcal{P}$; Forecasting scenarios set $\mathcal{S}=\{\text{Intersection, Roundabout, Merging}\}$; Training hyperparameter set $\{\mathcal{H}\}$; meta-model $m$; Total number of generations $\Theta$; Total number of sub-model $\Phi$.
\STATE Add meta-model $m$ into Knowledge Pool $\mathcal{P}$;
\FOR{$i \ in \ \{1,2,...\Theta\}$}
    \FOR {$s \in  \mathcal{S}$}
        \STATE Select meta-model from Knowledge Pool:  $m \sim \mathcal{P}$;
            \FOR {$j\ in \{1,2,...\Phi\}$}
            \STATE $m^{\prime} \gets \text{model\_evolutionary}(m)$;
            \STATE $m^{\prime} \gets \text{knowledge\_transfer}(m)$;
            \STATE $\{\mathcal{H}\}^{\prime} \gets \text{hyperparameter\_tunning}(\{\mathcal{H}\})$;
            \STATE Train the sub-model $m_i$ on scenario $s$ with hyperparameter set $\{\mathcal{H}\}^{\prime}$;
            \ENDFOR
        \STATE Add sub-model with the highest evaluation score into Knowledge Pool $\mathcal{P}$: \\ $\mathcal{P}\gets \mathcal{P}\bigcup \max(\text{evaluation\_score}(m^{\prime}))$;
    \ENDFOR
\ENDFOR
\end{algorithmic}
\end{algorithm}

    \begin{algorithm}[t]
    \caption{Parent Model Selection Algorithm}
    \begin{algorithmic}[1]
    \label{parent-select}
    \STATE \textbf{Input:} A scene set $\mathcal{S}$ to be trained; Number of sub-models $\mathcal{G}$ generated by each candidate model; Models set $\mathcal{M}$ from Knowledge Pool; List of model scores and ranks $\mathcal{R}$; Parent model $p$.
    \FOR {$m \in \mathcal{M}$}
        \STATE Evaluate $m$ on $\mathcal{S}$ to obtain its score $s_m$;
        \STATE Compute the rank $r_m$ of $m$: $r_m \gets s_m \times 0.9^{\mathcal{G}}$;
        \STATE Add $(m, r_m)$ into $\mathcal{R}$;
    \ENDFOR
    \STATE Sort $\mathcal{R}$ in descending order of rank;
    \STATE Set the highest ranked model in $\mathcal{R}$ as $p$.
    \end{algorithmic}
    \end{algorithm}

    \begin{figure}[ht!]
    \centering
   \includegraphics[width=0.9\linewidth]{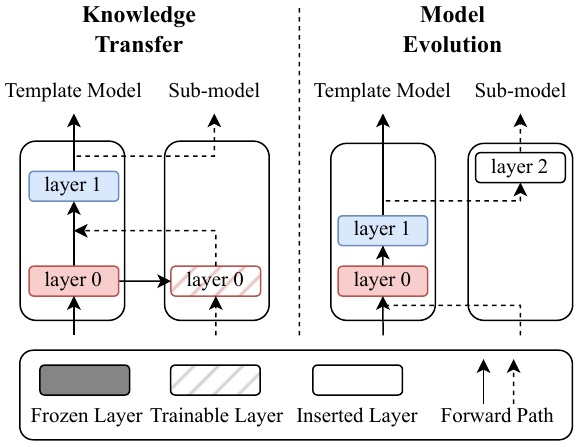}
    
   \caption{\textbf{Knowledge Transfer and Model Evolution in MS-Net training.} On the left, network parameters from the meta-model are inherited through a knowledge transfer process, solid layers are trainable in the sub-model, while hatched layers are non-trainable and reused. In the Model Evolution approach on the right, the sub-model adds a trainable layer (e.g., decoder) while reusing other frozen layers from the meta-model.
   }
   \label{fig2}
   
\end{figure}
 \subsection{Evolutionary Algorithm} 
    Inspired by genetic algorithms, we design an evolutionary algorithm to find the best model for the new scenario. The evolutionary algorithm enables knowledge to be partially inherited from existing models while increasing the knowledge according to new scenarios.

    The generation of sub-models from meta-models is based on the following rules:
    
    \textbf{Model Evolution}: Since the prediction difficulty varies from scenario to scenario, we designed the Model Evolution mechanism to dynamically compress or expand the motion forecasting models. 
    The motion forecasting model primarily depends on three distinct components: (1) trajectory encoding layers, which process historical trajectories; (2) map encoding layers, responsible for extracting road features from maps; and (3) the decoder layer, modeling interactions among agents and road context information. Our evolution algorithm only targets those components to perform mutational operations. Specifically, during the generation of sub-models, the last trajectory or map encoder layer and the last interaction decoder layer can be removed or a new layer can be appended, both with a probability defined by the \textit{mutation rate}. As a general observation, augmenting the number of layers in each component can enhance corresponding capacity, whereas reducing them might diminish its capabilities.

    Given a meta-model $m=\left(E_{\text {trajectory}}, E_{\text {map}}, D_{\text{interaction}}\right)$ where $E_{\text {trajectory }}=\left[e_{t1}, e_{t2}, \ldots, e_{L_{tn}}\right]$ is the trajectory encoder, $E_{\text {map }}=\left[e_{m1}, e_{m2}, \ldots, e_{mn}\right]$ is the map encoder $D_{\text{interaction}}=\left[d_1, d_2, \ldots, d_{n}\right]$ is the interaction decoder, let $\rho_{1}$ be the mutation rate for Model Evolution, and n be the number of layers in each component. For every component $C$ drawn from the set $\left\{E_{\text {trajectory }}, E_{\text {map }}, D_{\text {interaction }}\right\}$, the Model Evolution mutation is defined as follows:

    \begin{equation}
    C^{\prime}= \begin{cases}\operatorname{Insert}\left(C, c_{\text {i+1}}\right) & \text { if } \operatorname{rand}()<\rho_{1} / 2; \\ \operatorname{Delete}\left(C, c_i\right) & \text { if } \rho_{1}/ 2 \leq \operatorname{rand}()<\rho_{1}; \\ C & \text { otherwise },\end{cases}
    \end{equation}
    where $\operatorname{Insert}\left(C, c_{\text {i+1}}\right)$ adds a new layer $c_{\text {i+1}}$ at the end of component $C$. $\operatorname{Delete}\left(C, c_i\right)$ removes the last layer $c_i$ from model $m$. Thus, the resultant mutated sub-model $m^{\prime}$ will be: $m^{\prime}=\left(E_{\text {trajectory }}^{\prime}, E_{\text {map }}^{\prime}, D_{\text{interaction}}^{\prime}\right)$.
    
    The evolutionary result is illustrated using a transformer-based\cite{vaswani2017attention} model as an example shown in \cref{fig3}.

    \textbf{Knowledge Transfer}: To share common knowledge from different prediction scenes among the sub-models, the Knowledge Transfer mechanism randomly selects each layer $l$ in $m$ with a probability of \textit{mutation rate}. For each selected layer, the sub-model will create a copy from the parent model, as shown in \cref{fig2}(Left). Those copies of layers can be fine-tuned by the sub-model during the training process. To prevent catastrophic forgetting\cite{2017overcomingforgetting,hadsell2020embracing,hayes2020remind}, the remaining portion of the parent model layer is shared with the sub-model in a frozen state\cite{transferlearning}. Through knowledge transfer, the sub-model can leverage knowledge acquired from other scenes and optimize them for its own scene, thus enhancing its performance.
    For layer $l_i$ in meta-model $m$ , the layer $l_i^{\prime}$ in its sub-model $m^{\prime}$ is:
    \begin{align}
    l_i^{\prime} =~& \mathcal{I}(\operatorname{rand}()<\rho_{2}) \cdot \operatorname{Tune}\left(l_i\right)\cr
    &+ \mathcal{I}(\operatorname{rand}() \geq \rho_{2}) \cdot \operatorname{Froze}\left(l_i\right),
    \end{align}
    where $\operatorname{Tune}\left(l_i\right)$ create a replication of layer $l_I$ from the parent model and fine-tunes this layer within its specific scene, $\operatorname{Froze}\left(l_i\right)$ shared $l_i$ with its parent model with frozen state ensuring that the parameters remain static. $\mathcal{I}(\cdot)$ is an indicator function, which equals 1 if a condition is met and 0 otherwise. $\rho_{2}$ is the mutation rate for Knowledge Transfer.

    \textbf{Scoring Function}: The scoring function is utilized to select a sub-model $m_i$ from a pool of multiple candidate sub-models with a minimal number of additional parameters and superior accuracy.  As the sub-model's newly acquired knowledge is stored in trainable parameters, whereas the remaining parameters are directly inherited from the parent model, Occam's Razor dictates that the sub-model with the lowest number of additional parameters and highest accuracy should be chosen as the optimal sub-model. Our scoring matrix encourages the sub-model to share knowledge with others while retaining only the necessary scenario-distinct knowledge.

    \begin{equation}
    \operatorname{score}(m_i)=q(m_i) * a ^{p(m_i)}, a \in[0,1],
    \end{equation}
    where $p(m_i)$ is the additional parameters in sub-model $m_i$, $q(m_i)$ is its quality (e.g., average distance error, miss rate) from the validation set, and $a$ is the new parameters' penalty factor.

    \textbf{Hyperparameter Tuning}: Due to the prediction difficulty varying across different scenarios, the hyperparameters for the training procedure of the sub-model are not optimized for each unique scenario initially. To address this challenge, we adopt the random walk algorithm to tune these hyperparameters, thereby improving the generalization performance of the sub-models. Specifically, for the optimizer \cite{zhou2020pbsgd} hyperparameter(e.g. learning rate, weight decay), it will be randomly selected with a probability of \textit{mutation rate}, and then slightly adjusted around its previous value using the random walk algorithm. 

    Given a candidate set for one hyperparameter $h$ represented as: $\hat{\mathcal{H}}=\left[h_1, h_2, \ldots, h_N\right]$. For a previous value $h_i$, its mutated value $h_i^{\prime}$ can be given by:
    \begin{equation}
    h_i^{\prime}= \begin{cases}\operatorname{\hat{\mathcal{H}}}[i-1] & \text { if } \operatorname{rand}()<\frac{\rho}{2} \text { and } i>1; \\ \operatorname{\hat{\mathcal{H}}}[i+1] & \text { if } \frac{\rho}{2} \leq \operatorname{rand}()<\rho \text { and } i<N; \\ h_i & \text { otherwise }.\end{cases}
    \end{equation}

\section{Experiment}

\subsection{Experimental Setup}
    
    \textbf{Datasets}: In this study, we evaluated the effectiveness of our proposed method using the ETH\cite{21}, UCY\cite{22}, and INTERACTION datasets\cite{INTERACTIONDATASET}. All datasets are divided into multiple scenes, making them suitable for our experimental purposes. The ETH/UCY datasets contain 2206 pedestrian trajectories from settings like university campuses, train stations, and malls. We are required to predict future trajectories of 12 timesteps (4.8s) using observed trajectories from the past 8 timesteps (3.2s). To ensure fair comparisons, we did not use any semantic or visual information for the ETH/UCY datasets, similar to the baseline settings\cite{predictionTransfomer, AgentFormer}. The INTERACTION dataset contains 40,054 records of vehicles collected from intersections, mergers, and roundabouts. We are required to forecast the next 3 seconds based on 1 second of past observations, sampled at 10 Hz.
    
    \textbf{Metrics}: We use Average Displacement Error: 
    \begin{equation}
    \operatorname{ADE}_K=\frac{1}{T} \min _{k=1}^K \sum_{t=1}^T\left\|\hat{\mathbf{y}}_n^{t,(k)}-\mathbf{y}_n^t\right\|^2,
    \end{equation}
    which measures the overall closeness of the predicted results to the true value, and averaging the discrepancy at each time step. $\hat{\mathbf{y}}_n^{t(k)}$ denotes the future position of agent $n$ at time $t$ in the $k$-th sample and $\mathbf{y}_n^T$ is the corresponding ground truth. $T$ is the number of predicted timestamps and $K$ is the number of modalities.

    To evaluate the accuracy of the prediction at the last time step, we use the Final Displacement Error: 
    \begin{equation}
    \operatorname{FDE}_K=\frac{1}{T} \min _{k=1}^K\left\|\hat{\mathbf{y}}_n^{T,(k)}-\mathbf{y}_n^T\right\|^2,
    \end{equation}
    ADE and FDE are the standard metrics for trajectory prediction \cite{21,22}.  
    Minimum Joint Average Displacement Error (minJointADE) is set as the metric by the official guidance of the INTERACTION Prediction challenge, which aims to evaluate the model’s joint prediction performance under highly interactive scenarios. minJointADE represents the minimum value of the Euclid distance averaged by time and all agents between the ground truth and modality with the lowest value. It can be calculated by 
\begin{equation}
    \operatorname{minJointADE}=\min _{k \in\{1, \ldots, K\}} \frac{1}{N T} \sum_{n, t} \sqrt{\left(\hat{\mathbf{y}}_{n}^{t,k}-\mathbf{y}_n^t\right)^2},
\end{equation}
where $N$ is the number of agents to be predicted in this case.

\textbf{Implementation Details}: We implemented our model using Pytorch and conducted all experiments on an NVIDIA Tesla V100. The mutation rate was set to 0.2, $a$ for the evaluation function was set to 0.8, and MS-Net was trained for 3 generations sampling three sub-models for each scenario within the ETH/UCY dataset and five sub-models for the INTERACTION dataset. All other hyperparameters followed the original settings of the meta-model. For the ETH/UCY datasets, we treated each of the eight training sets as a distinct scenario and learned a forward passing path for each one, while the baseline method used all eight training sets. For the INTERACTION dataset, we adopted its pre-existing scene division method and trained a separate path of our network for each of the three scenes, i.e., intersections, merging, and roundabouts, to perform trajectory prediction.

\begin{table*}[ht!]
  \caption{Ablation studies on ETH/UCY datasets. We use the TF\cite{predictionTransfomer} model as the meta-model.}
  \centering
  \begin{tabular}{cccccc}
    \toprule
    \multicolumn{3}{c}{\textbf{Evolutionary Algorithm}} & \multicolumn{2}{c}{\textbf{Accuracy}} & \textbf{Model Size ($\times 10^6$)} \\ 
    \cmidrule(ll){1-3} \cmidrule(ll){4-5} \cmidrule(ll){6-6}
    \multicolumn{1}{c}{Hyperparameter Tuning} & \multicolumn{1}{c}{Evolutionary Model} & \multicolumn{1}{c}{Evaluation Function} & Avg ADE$\downarrow$ & \multicolumn{1}{c}{Avg FDE$\downarrow$} & Average Effective Parameters$\downarrow$ \\ 
    \midrule
    \checkmark & & & 0.544 & 1.127    & 29.000    \\
    & \checkmark &    &0.494 & 1.049   & 16.400    \\
    &  & \checkmark  & 0.517  & 1.097  & /  \\
    \checkmark  & & \checkmark &  0.499  & 1.045   & 20.180 \\
    \checkmark  & \checkmark & & 0.490  & 1.016   & 10.500 \\
    & \checkmark   & \checkmark & 0.575 & 1.179 & 3.580  \\
    \checkmark & \checkmark  & \checkmark    & 0.495 & 1.040  & 4.890 \\                              
    
    \bottomrule
  \end{tabular}
  \label{tab:ablation}
\end{table*}
\subsection{Comparison Experiment}
	We employed the TF\cite{predictionTransfomer}, Agentformer\cite{AgentFormer}, and AutoBot\cite{autobot} models as the meta-model separately to prove the effectiveness of our method. As shown in \cref{tab1}, The average Displacement Error of MS-Net based on a standard Transformer decreases by 9\%, and the Final Displacement Error decreases by 13.4\%. The TF model used in the original paper includes six layers of encoders and six layers of decoders. As shown \cref{fig3}, our network uses only one-third of the parameters, reduces inference time by 25\%, and uses 35\% less GPU memory. Despite this, it showed better prediction accuracy than the TF model.  

    Additionally, we incorporated Agentformer as our meta-model. Agentformer is a leading pedestrian trajectory prediction model that does not require map information. Experiments demonstrate that our algorithm significantly improves the meta-model's performance, reducing the Average Displacement Error by an average of 14\% and the Final Displacement Error by 22.4\%, as illustrated in \cref{tab2}. As MS-Net's forward inference computation remains the same as the original model, as no model evolution was performed, the improvement is notable.

\begin{figure}[ht!]
\centering
\includegraphics[width=0.9\linewidth]{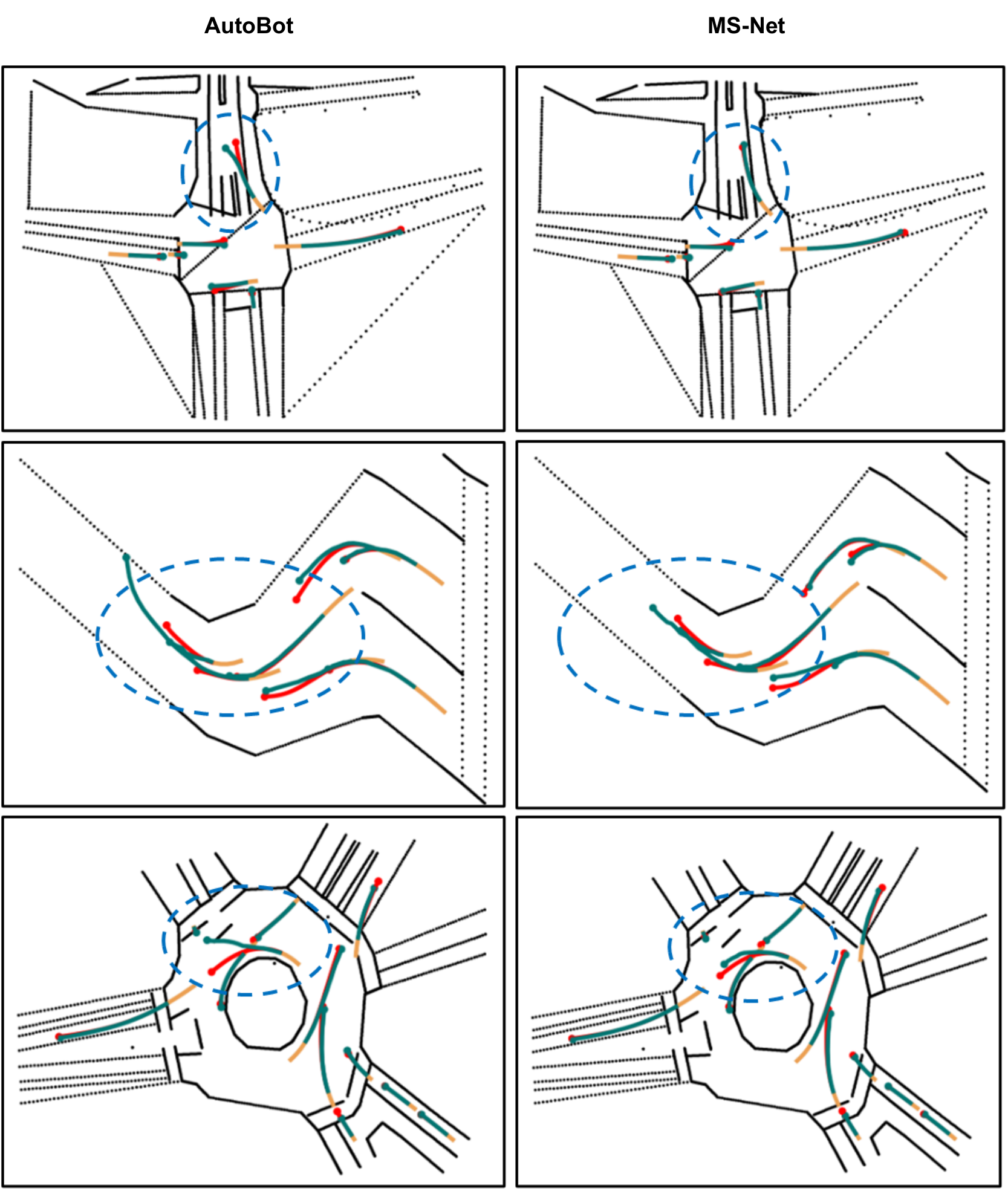}
\caption{\textbf{Comparation of AutoBot\cite{autobot} with MS-Net on INTERACTION validation set.} The past trajectories are shown in yellow, the ground-truth trajectories are shown in red, and the predicted trajectories are shown in green.}
\label{quantitiveresult}
\end{figure}

    To validate the versatility of our approach, we assessed our method on the INTERACTION dataset, utilizing AutoBot as our meta-model. As depicted in \cref{tab:MS-Autobot}, MS-Net ranked second on the INTERACTION dataset leaderboard. Our qualitative results, as illustrated in \cref{quantitiveresult}, demonstrate that MS-Net exhibits superior performance when confronted with challenging prediction instances across various scenarios. We attribute this to MS-Net's sparsity, which allows it to possess more scenario-distinct knowledge, enabling it to handle these hard cases better than the Unified model. 

    We compared MS-Net with other optimization methods on the INTERACTION dataset (\cref{tab:evo-other-tune-methon}). The Parallel Model, using three separate models, had the worst results due to a lack of shared knowledge across scenes. Multi-Head sharing a model backbone while fine-tuning the final fully-connected output layer on each scene which alleviates the knowledge-sharing setback. The Unified Model employs a single model trained with complete data from all three scenes, serving as a baseline. In comparison, MS-Net achieved superior prediction accuracy while maintaining a small total parameter number compared to the Parallel Model and maintaining the same inference computation load as the Unified Model. MS-Net (Model-evo) allows dynamic insertion or removal of layers from the meta-model achieving comparable performance while enhancing efficiency by reducing inference time by 20.8\% and lowering GPU memory usage by 27.3\%. 
    
\begin{table}[ht]
  \centering
  \caption{The performance of MS-Net with Transformer as the meta-model on the test set, K = 20 Samples.}
  \begin{tabular}{ccccc}
    \toprule
    \multirow{2}{*}{\textbf{Scenario}} & \multicolumn{2}{c}{\textbf{Transformer}\cite{predictionTransfomer}} & \multicolumn{2}{c}{\textbf{MS-Net}} \\ 
    \cmidrule(ll){2-3} \cmidrule(ll){4-5}
    ~ &  ADE$\downarrow$      & FDE$\downarrow$       & ADE$\downarrow$     & FDE$\downarrow$    \\
    \midrule
    ETH & 1.030 & 2.100 & \textbf{0.995 (-3.4\%)} & \textbf{2.045 (-2.6\%)} \\
    HOTEL & 0.360 & 0.710 & \textbf{0.259 (-28.5\%)} & \textbf{0.470 (-33.8\%)} \\
    UCY & 0.530 & 1.320 & \textbf{0.512 (-3.4\%)} & \textbf{1.101 (-16.6\%)} \\
    ZARA1 & 0.440 & 1.000 & \textbf{0.434 (-1.5\%)} & \textbf{0.958 (-4.2\%)} \\
    ZARA2 & 0.340 & 0.760 & \textbf{0.312 (-8.2\%)} & \textbf{0.686 (-9.8\%)} \\
    \bottomrule
  \end{tabular}
  \label{tab1}
\end{table}

\begin{table}[ht]
  \centering
  \caption{The performance of MS-Net with AgentFormer as the meta-model on the test set, K = 20 Samples.}
  \begin{tabular}{ccccc}
    \toprule
    \multirow{2}{*}{\textbf{Scenario}} &  \multicolumn{2}{c}{\textbf{AgentFormer}\cite{autobot}} & \multicolumn{2}{c}{\textbf{MS-Net}} \\ 
    \cmidrule(ll){2-3} \cmidrule(ll){4-5}
    ~ &  ADE$\downarrow$      & FDE$\downarrow$       & ADE$\downarrow$     & FDE$\downarrow$    \\
    \midrule
ETH & 0.442 & 0.720 & \textbf{0.436 (-1.5\%)} & \textbf{0.699 (-2.9\%)} \\
HOTEL & 0.295 & 0.561 & \textbf{0.204 (-30.9\%)} & \textbf{0.331 (-41.1\%)} \\
UCY & 0.369 & 0.668 & \textbf{0.338 (-8.3\%)} & \textbf{0.596 (-10.8\%)} \\
ZARA1 & 0.419 & 0.890 & \textbf{0.340 (-18.8\%)} & \textbf{0.637 (-28.4\%)} \\
ZARA2 & 0.284 & 0.579 & \textbf{0.246 (-13.2\%)} & \textbf{0.410 (-29.2\%)} \\
    \bottomrule
  \end{tabular}
  \label{tab2}
\end{table}

\begin{table}[ht]
    \caption{Comparison results of the test set from the INTERACTION challenge}

    \begin{tabular}{cccc}
    \toprule
    \textbf{Method}                         & \textbf{MinJointADE$\downarrow$} & \textbf{MinJointFDE$\downarrow$} & \textbf{MinJointMR$\downarrow$} \\ 
    \midrule
    \multicolumn{1}{c}{ReCoG\cite{mo2020recog}}    & 0.4668     & 1.1597     & 0.2377    \\
    \multicolumn{1}{c}{DenseTNT\cite{Gu2021DenseTNTET}}  & 0.4195     & 1.1288     & 0.2240    \\
    \multicolumn{1}{c}{AutoBot\cite{krlj2021autoBOTEN}}   & 0.3123     & 1.0148     & 0.1933    \\
    \multicolumn{1}{c}{Traj-MAE\cite{Chen2023TrajMAEMA}}  & 0.3066     & 0.9660     & 0.1831    \\
    \multicolumn{1}{c}{FJMP (1st)\cite{Rowe2022FJMPFJ}} & 0.2771     & 0.9170     & 0.1742    \\
    \multicolumn{1}{c}{\textbf{Ours (2nd)}} & 0.3003     & 0.9619     & 0.1832    \\ 
    \bottomrule
    \end{tabular}
\label{tab:MS-Autobot}
\end{table}

\begin{table}[ht]
\caption{Results compare on INTERACTION dataset Validation dataset}

\begin{tabular}{cccc}
\toprule
\textbf{Method}                  & \textbf{\makecell[c]{Average\\ minADE$\downarrow$}} & \textbf{\makecell[c]{Inference\\ params \\($\times 10^6$)}}  & \textbf{\makecell[c]{Total\\params \\($\times 10^6$)}}  \\ 
\midrule
Parallel Model          & 0.2532          & 2.71          & 2.71*3       \\
Multi-Head              & 0.2279          & 2.71          & 0.03*3+2.68 \\
Unified Model (AutoBot) & 0.2045          & 2.71          & 2.71         \\
MS-Net                  & \textbf{0.2015} & 2.71          & 6.90         \\
MS-Net (Model-Evo)       & 0.2471          & \textbf{1.78} & 5.33         \\ 
\bottomrule
\end{tabular}
\label{tab:evo-other-tune-methon}
\end{table}

\subsection{Ablation Study}
    We performed ablation studies on the key components of our algorithm, including Hyperparameter Tuning, Evolutionary Model, and Evaluation Function. To demonstrate the effectiveness of the Evaluation Function, we incorporated the Average Accounted Parameter Numbers as part of our metrics. Notably, in instances where we did not execute the Evolutionary Model, we manually experimented with various parameter values and recorded the highest accuracy achieved.
    
    As shown in \cref{tab:ablation}, each part of the algorithm plays an important role in our approach. Without an Evaluation function, which means that the highest prediction accuracy can be achieved when we do not limit the number of parameters in the model.  
    Hyperparameter Tuning enables us to attain superior performance in new scenarios. Changes in hyperparameters impact the training process and performance scores of child models, subsequently influencing the evolutionary paths chosen in the construction of the final model. By adaptively cropping or expanding the model, the network can adjust the number of parameters to the scene's complexity, but this often results in using a more complex model. The Evaluation Function provides an effective solution. The Evaluation Function balances the model's accuracy against the number of new parameters introduced, obtaining a lightweight model with comparable accuracy while avoiding overfitting the dataset.

\section{Discussion}

    MS-Net's training time is approximately $1.3\times$ that of TF\cite{predictionTransfomer} and $3\times$ of AutoBot\cite{autobot} due to its sampling of multiple child models. However, it uniquely activates a single path per scenario. On the UCY/ETH dataset, MS-Net is 25\% faster than TF, using 35\% less GPU memory. On the INTERACTION dataset, it's 20.8\% faster than AutoBot, consuming 27.3\% less GPU memory.

    The performance difference between MS-Net and FJMP\cite{Rowe2022FJMPFJ} stems from FJMP's use of a directed acyclic graph, providing detailed agent interactions, beneficial for datasets like INTERACTION. In contrast, AutoBot\cite{autobot} employs a connected graph within a transformer, which is less adept at capturing complex interactions.

\section{Conclusion}
In this paper, we introduce the MS-Net, a novel approach designed for the task of multi-scene motion prediction in autonomous driving systems. The MS-Net employs an adaptive strategy, which involves the evolution of sub-models based on meta-models. This adaptability is rooted in the dynamic management of Model Evolution, Knowledge Transfer, and Evaluation Functions during the training phase, allowing it to effectively respond to scene diversity.
Our extensive evaluations on the ETH/UCY and INTERACTION datasets highlight the strong performance of MS-Net. It matches or surpasses existing methods in accuracy while significantly reducing computational costs by 33$\%$ and 64$\%$ during inference, on ETH/UCY and INTERACTION respectively. This advantage is of great importance in autonomous driving, which needs swift decision-making with constrained computational resources.
Looking ahead, we see potential in integrating reinforcement learning to enhance evaluation processes, aiming for improved prediction precision. Additionally, applying MS-Net across various datasets to facilitate knowledge transfer between different domains is a direction worth exploring.

{\Large
\bibliographystyle{IEEEtran}
\bibliography{main}
}

\vfill

\end{document}